\pdfoutput=1

\documentclass[11pt]{article}

\usepackage[preprint]{acl}
\usepackage{tikz}
\usepackage{times}
\usepackage{latexsym}
\usepackage{xcolor}
\usepackage{amsmath}
\usepackage{amssymb}
\usepackage{booktabs}
\usepackage[T1]{fontenc}

\usepackage[utf8]{inputenc}

\usepackage{microtype}

\usepackage{inconsolata}

\usepackage{graphicx}

\title{Reducing Hallucinations in Summarization via Reinforcement Learning with Entity Hallucination Index}

\author{
  \textbf{Praveenkumar Katwe\textsuperscript{1,2}},
  \textbf{Rakesh Chandra Balabantaray \textsuperscript{1}},
  \textbf{Kali Prasad Vittala\textsuperscript{2}}\\
  \textsuperscript{1}International Institute of Information Technology,Bhubaneshwar,INDIA,\\
  \textsuperscript{2}Informatica Business Solutions, Bengaluru , INDIA \\
 \small{
    \textbf{Correspondence:} {c121007@iiit-bh.ac.in},{rakesh@iiit-bh.ac.in},{kprasad@informatica.com} 
}
}

\begin{document}
\maketitle
\begin{abstract}
Reducing hallucinations in abstractive summarization remains a critical challenge for deploying language models (LMs) in real-world settings. In this work, we introduce a reward-driven fine-tuning framework that explicitly optimizes for Entity Hallucination Index (EHI), a metric designed to quantify the presence, correctness, and grounding of named entities in generated summaries. Given a corpus of meeting transcripts, we first generate baseline summaries using a pre-trained LM and compute EHI scores via automatic entity extraction and matching. We then apply reinforcement learning to fine-tune the model parameters, using EHI as a reward signal to bias generation toward entity-faithful outputs. Our approach does not rely on human-written factuality annotations, enabling scalable fine-tuning. Experiments demonstrate consistent improvements in EHI across datasets, with qualitative analysis revealing a significant reduction in entity-level hallucinations without degradation in fluency or informativeness. We release a reproducible Colab pipeline, facilitating further research on hallucination-aware model fine-tuning using lightweight, hallucintion metrics like EHI.
\end{abstract}

\section{Introduction}

Abstractive summarization models, powered by large language models (LLMs), have achieved impressive results across various domains. However, a persistent challenge remains: \textbf{hallucination}, where generated summaries include incorrect or fabricated information not grounded in the source input \citep{maynez2020faithfulness, durmus2020feqa}. In high-stakes applications such as meeting summarization, medical reporting, or financial documentation, hallucinations, particularly involving named entities, can significantly undermine trustworthiness and utility.

Existing work on hallucination detection primarily focuses on coarse-grained factuality metrics or relies on reference-based evaluations \citep{falke2019ranking, pagnoni2021understanding}. While such methods are valuable, they often fail to capture fine-grained entity-level hallucinations, and their dependence on ground-truth references limits scalability. Recent efforts have explored lightweight automatic metrics, yet integrating such evaluations directly into model training remains underexplored.

In this work, we propose a novel fine-tuning approach that leverages the \textbf{Entity Hallucination Index (EHI)} as a reward signal to reduce hallucinations at the entity level. EHI quantifies the correctness and grounding of entities by comparing extracted entities from model outputs to those present in the input document, enabling efficient  evaluation with reduced reference. We first generate baseline summaries using a pre-trained model, compute EHI scores through automatic entity extraction and matching, and fine-tune the model using reinforcement learning with EHI as the reward function. This method biases the model toward producing more entity-faithful summaries without requiring human factuality annotations.

Our contributions are as follows:
\begin{itemize}
    \item We introduce an \textbf{EHI-guided fine-tuning framework} that improves entity faithfulness in abstractive summarization.
    \item We present a \textbf{scalable reinforcement learning pipeline} that does not depend on human-labeled factuality datasets.
    \item We demonstrate \textbf{empirical improvements} in EHI scores across meeting transcript datasets and conduct qualitative analysis to highlight reductions in hallucination.
    \item We release a \textbf{reproducible Colab-based implementation} to facilitate further research on hallucination-aware summarization.
\end{itemize}

Our findings suggest that optimizing language models with fine-grained entity-focused rewards like EHI provides an effective pathway for enhancing faithfulness and reliability in summarization tasks.

\section{Related Work}

\paragraph{Hallucination in Summarization.}
Hallucination, or the generation of content that is unfaithful to the input, is a long-standing challenge in abstractive summarization \citep{maynez2020faithfulness}. Several studies have quantified hallucination prevalence across datasets and models \citep{pagnoni2021understanding}. Techniques such as constrained decoding \citep{cao2020factual} and fact-checking-based reranking \citep{goyal2022fact} have been proposed to mitigate hallucination during or after generation.

\paragraph{Entity-Level Hallucination Detection.}
Prior work has recognized the importance of entity-level evaluation for factuality. Metrics like FEQA \citep{durmus2020feqa} and QuestEval \citep{scialom2021questeval} attempt to assess factual consistency using question answering techniques. However, these methods often require reference summaries or external QA systems, limiting scalability. Our work builds on this line by employing the \textbf{Entity Hallucination Index (EHI)}, a lightweight evaluation metric focusing specifically on named entity faithfulness.

\paragraph{Reward-Driven Fine-Tuning.}
Reinforcement learning from human feedback (RLHF) has gained traction for aligning model outputs with desired qualities, such as factuality and helpfulness \citep{ouyang2022training}. In summarization, factuality-aware reward models have been used for fine-tuning \citep{pasunuru2018multi}. Unlike previous approaches that rely on coarse factuality rewards or human-labeled datasets, we propose fine-tuning using EHI as a direct, automatic reward signal, thereby promoting entity-grounded summarization without additional supervision.

\section{Methodology}

\begin{figure*}[t]
\centering
\begin{tikzpicture}[node distance=2cm, auto]
\tikzstyle{block} = [rectangle, draw, fill=blue!20, 
    text width=8em, text centered, rounded corners, minimum height=3em]
\tikzstyle{line} = [draw, -latex]

\node [block] (input) {Input Transcript};
\node [block, right of=input, node distance=4cm] (summarizer) {Baseline Summarization Model};
\node [block, right of=summarizer, node distance=5cm] (ehi) {EHI Scoring};
\node [block, below of=ehi, node distance=2.5cm] (finetune) {Reward-Based Fine-Tuning Update};
\path [line] (input) -- (summarizer);
\path [line] (summarizer) -- (ehi);
\path [line] (ehi) -- (finetune);
\path [line] (finetune.west) -- ++(-5cm,0) -- (input.south);

\end{tikzpicture}
\caption{Overview of our fine-tuning pipeline using EHI rewards.}
\label{fig:pipeline}
\end{figure*}
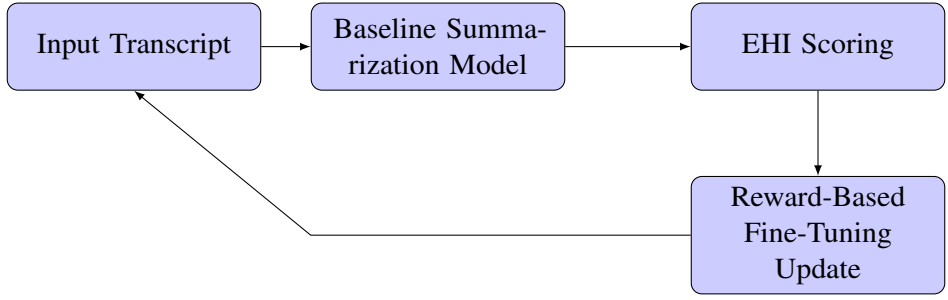

Our objective is to fine-tune a language model (LM) to generate summaries with reduced entity hallucination. We achieve this by designing a reward function based on the Entity Hallucination Index (EHI) and applying reinforcement learning to maximize it. Figure~\ref{fig:pipeline} presents an overview of our pipeline.

\subsection{Problem Formulation}
Given an input document $x$, the goal is to generate a summary $y$ such that entities mentioned in $y$ are grounded in $x$. Traditional maximum likelihood training does not explicitly optimize for factual consistency. Instead, we introduce a reward-driven fine-tuning strategy where the reward is proportional to entity faithfulness, measured using EHI.
\subsection{Datasets Preparations}
We use the official datasets provided in the AutoMin 2025 shared task \footnote{\url{https://ufal.github.io/automin-2025/}}, which includes the ELITR Minuting Corpus for summarization and the ELITR-Bench dataset for question answering \footnote{\url{https://github.com/utter-project/ELITR-Bench/}}. We followed the task protocols as outlined in the shared task instructions, including the official train/dev/test splits. For summarization (Task A), gold summaries are provided for reference-based evaluation. For question answering (Task B), answer spans are matched against gold annotations for exact match and F1 scoring.
We evaluate our approach on given meeting transcript datasets, which consist of multi-turn conversational dialogues along with abstractive gold summaries. Each sample is formatted as a flattened transcript-summary pair. To simulate real-world entity usage, datasets include a diverse range of named entities spanning persons, organizations, locations, and events. We randomly partition each dataset into training (80\%), validation (10\%), and test (10\%) splits.

Entity extraction is performed on both source documents and summaries for computing the Entity Hallucination Index (EHI). During fine-tuning, we exclusively utilize the training split, while validation EHI scores are monitored to select the best checkpoint. Final evaluations are reported on the held-out test set.

\begin{figure*}[t]
\includegraphics[width=1\textwidth]{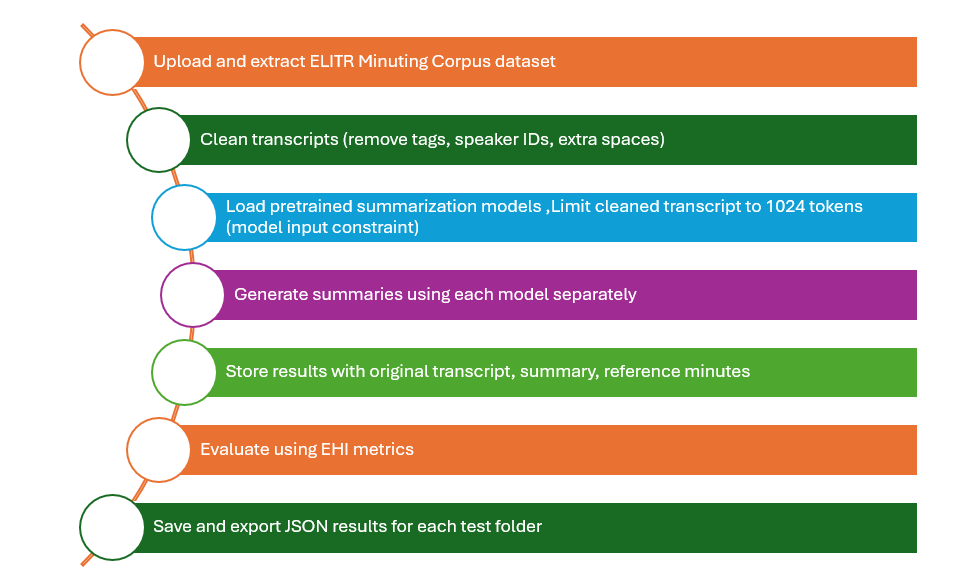}
\caption{TASK B: Question-Answer Generation}
\label{fig:Task B process flow}
\end{figure*}

\subsection{Baseline Summarization}
We begin with a pre-trained encoder-decoder LM (e.g., \texttt{Flan-T5-Large}) fine-tuned on generic summarization tasks. For each input document, the model generates an initial summary using standard greedy or beam search decoding strategies.

\subsection{Entity Hallucination Index Computation}
To assess entity faithfulness, We formally define the \textbf{Entity Hallucination Index (EHI)} as:

\begin{equation}
\text{EHI} = \frac{e^{\text{PH}} + e^{\text{EF}}}{e^{\text{PH}} + e^{\text{EF}} + e^{\text{NH}} + e^{\text{OF}} + e^{\text{LF}}}
\end{equation}
where \( x_i \in \{ \text{PH}, \text{EF}, \text{NH}, \text{OF}, \text{LF} \} \) represent scores corresponding to Positive Hallucination (PH), Extractiveness Factor (EF), Negative Hallucination (NH), Overfocused Relations (OF), and Lost Focus (LF), respectively.
We define the following factors:
\begin{itemize}
    \item \textbf{Positive Hallucination (PH):} Measures newly introduced entities that are factually correct and beneficial.
    \item \textbf{Extractiveness Factor (EF):} Measures entities accurately extracted from the input document into the summary.
    \item \textbf{Negative Hallucination (NH):} Captures hallucinated entities that are incorrect or not grounded in the input.
    \item \textbf{Overfocused Relations (OF):} Penalizes summaries that overly focus on a narrow subset of entities, missing diversity.
    \item \textbf{Lost Focus (LF):} Penalizes summaries that omit important entities present in the input.
\end{itemize}

Higher values of EHI indicate better entity faithfulness. While the term ``Entity Hallucination Index'' may suggest a higher hallucination rate, our formulation is designed such that EHI rewards desirable behaviors — namely, correctly extracted entities (EF) and factually accurate, beneficial additions (PH) — while penalizing undesirable behaviors such as ungrounded entities (NH), overfocus (OF), and loss of important content (LF). 

This index behaves more like a reward function, where a higher score means fewer harmful hallucinations and more helpful entity alignment. We retain the name EHI for continuity but note that it acts effectively as a precision-weighted reward.
Note that when all error factors (NH, OF, LF) are zero, the EHI reduces to 1, regardless of PH and EF values. This behavior reflects that the summary is fully faithful, and additional correct entities (PH) or extractions (EF) do not reduce the score.

Entity extraction is performed using a named entity recognition (NER) model (\texttt{en\_core\_web\_sm} from spaCy). Matching is case-insensitive and performed at the entity string level. As described in Section 3.4, higher EHI scores reflect better entity alignment and reduced hallucination indicating more entity-faithful summaries.

\begin{table*}[t]
\centering
\begin{tabular}{ll}
\hline
\textbf{Hyperparameter} & \textbf{Value} \\
\hline
Model & Flan-T5-Large \\
Optimizer & Adam \\
Learning Rate & $5 \times 10^{-6}$ \\
Batch Size & Dynamic (based on 40GB GPU) \\
Sampling Strategy & Greedy Decoding \\
Reward Normalization & Applied per batch \\
Summary Regeneration & Every 500 updates \\
NER Tool & spaCy (\texttt{en\_core\_web\_sm}) \\
\hline
\end{tabular}
\caption{Hyperparameters and settings used during fine-tuning.}
\label{tab:hyperparams}
\end{table*}

\subsection{Reward-Driven Fine-Tuning}

Our objective is to fine-tune the model parameters $\theta$ by maximizing the expected Entity Hallucination Index (EHI) reward across generated summaries. Formally, for an input document $x$, a generated summary $y$, and model parameters $\theta$, the expected reward objective is defined as:

\begin{equation}
J(\theta) = \mathbb{E}_{y \sim p_\theta(y|x)} [\text{EHI}(y, x)]
\end{equation}

where $\text{EHI}(y, x)$ denotes the Entity Hallucination Index computed between the generated summary $y$ and the input $x$.

Following the REINFORCE algorithm \citep{williams1992simple}, the gradient of this objective is estimated as:

\begin{equation}
\nabla_\theta J(\theta) = \mathbb{E}_{y \sim p_\theta(y|x)} [\text{EHI}(y, x) \nabla_\theta \log p_\theta(y|x)]
\end{equation}

In practice, we sample summaries from the model, compute their corresponding EHI scores, and apply policy gradient updates to encourage higher-reward outputs. To stabilize training, we normalize EHI rewards within each batch and use the Adam optimizer with a small learning rate (e.g., $5 \times 10^{-6}$). Summaries are regenerated periodically to reflect model improvements during fine-tuning.
We directly fine-tune the pre-trained model using reinforcement learning, without any intermediate supervised cross-entropy training. The training objective is to maximize the Entity Hallucination Index (EHI) reward. Following the REINFORCE algorithm \citep{williams1992simple}, we estimate gradients based on sampled summaries and use the Adam optimizer to update model parameters. EHI rewards are normalized within each training batch to ensure stable learning.

\subsection{Evaluation Metrics}

\begin{table*}[h]
\centering
\begin{tabular}{lcc}
\hline
\textbf{Metric} & \textbf{Before Fine-Tuning} & \textbf{After Fine-Tuning} \\
\hline
EHI Range & 0.0 – 0.55+ & 0.3 – 0.6 \\
EHI Behavior & Volatile, inconsistent & \textcolor{green!50!black}{Stable}, moderate \\
Entity F1 Score & Mostly < 0.5, many zeros & Frequently 0.6–1.0 \\
F1 Behavior & Irregular, poor precision & High accuracy and \textcolor{green!50!black}{consistency} \\
EHI–F1 Correlation & Weak/unclear & Strong inverse correlation \\
\hline
\end{tabular}
\caption{Comparison of EHI and Entity F1 before and after reinforcement learning on meeting transcripts. 
\textcolor{green!50!black}{Green-marked improvements} are attributed to reward normalization and convergence effects of the Adam optimizer.}
\label{tab:results}
\end{table*}

We evaluate model performance primarily using the \textbf{Entity Hallucination Index (EHI)}. For complementary analysis, we also report \textbf{Entity F1} scores, computed by matching extracted named entities between the generated summary and the reference summary. This provides an additional measure of summary completeness and precision at the entity level.
Formally, let $E_{ref}$ be the set of named entities in the reference summary and $E_{gen}$ be the set of entities in the generated summary. Then:

\begin{equation}
\text{Precision} =\frac{|E_{ref} \cap E_{gen}|}{|E_{gen}|} 
\end{equation} \\
\begin{equation}
\text{Recall} =\frac{|E_{ref} \cap E_{gen}|}{|E_{ref}|} 
\end{equation}
\begin{equation}
\text{Entity F1} = 2 \cdot \frac{\text{Precision} \cdot \text{Recall}}{\text{Precision} + \text{Recall}}
\end{equation}

All evaluations are conducted on the held-out test split.
As part of Automin 2025, the content was available on evaluation for two task. 
\subsubsection{Evaluation Metrics for Task A}
The system output has been evaluated by Automin 2025.ROUGE Scores:
ROUGE (Recall-Oriented Understudy for Gisting Evaluation) measures the overlap of textual units like n-grams between a generated text and one or more reference texts.These metrics usually report precision, recall, and their harmonic mean (F1-score) as evaluation scores.\cite{lin-2004-rouge}
ROUGE-1 (R1): Measures overlap of unigrams (single words) between the candidate and reference text.
ROUGE-2 (R2): Measures overlap of bigrams (two-word sequences).
ROUGE-L (RL): Measures the longest common subsequence (LCS) between the candidate and reference, capturing sentence-level structure and fluency.
BERTScore:An advanced evaluation metric that uses BERT's contextual embeddings to measure semantic similarity between candidate and reference texts.Instead of exact token overlap, it compares token embeddings using cosine similarity, accounting for context and nuanced meaning.It calculates precision, recall, and F1 by aligning tokens from candidate and reference based on embedding similarity, thus better capturing paraphrasing and synonymy.BERTScore correlates better with human judgments compared to traditional metrics like ROUGE and BLEU.\cite{Zhang*2020BERTScore:}
BARTScore:A newer evaluation metric leveraging the BART model, a pretrained sequence-to-sequence transformer.It evaluates the likelihood of the generated text conditioned on the reference (or vice versa), providing a quality score based on model-based scoring rather than surface overlap.This method is particularly suited for text generation tasks like summarization and translation and has shown promising correlation with human evaluation \cite{10.5555/3540261.3542349}
\subsubsection{Evaluation Metrics for Task B}
The evaluation has been conducted using GPT-4o as the LLM-judge with a maximum score of 10, following the methodology detailed in the ELITR-Bench paper \cite{thonet2024elitrbench} and a mean score was provided for the responses. 

\subsection{Implementation Details}

We initialize our experiments using the \texttt{Flan-T5-Large} model \citep{chung2022scaling} as the base summarization system. Fine-tuning is conducted with the Adam optimizer \citep{kingma2015adam} and a learning rate of $5 \times 10^{-6}$. We train on meeting transcript datasets formatted as input-summary pairs, with batch sizes adjusted to fit a single NVIDIA A100 GPU (40GB memory).
Since Flan-T5-Large has a limited context window of 1024 tokens, we applied a chunking strategy to handle long meeting transcripts (often exceeding 10,000 tokens). Transcripts were split into overlapping chunks of up to 950 tokens with a stride of 200 tokens, ensuring contextual continuity. Each chunk was summarized independently, and final summaries were aggregated. This approach ensured compatibility with the model while preserving content fidelity across segments.

Summaries are generated using greedy decoding during both pre-fine-tuning and intermediate stages of training. EHI scores are computed by extracting named entities with spaCy's \texttt{en\_core\_web\_sm} model \citep{spacy2}, and matching is performed case-insensitively.

To ensure stability, we normalize EHI rewards within each training batch. Summaries are regenerated periodically every $500$ updates to reflect model improvements in reward estimation. Fine-tuning is continued until EHI scores converge on a held-out validation set.

\begin{figure*}[t]
\centering
\includegraphics[width=0.45\textwidth]{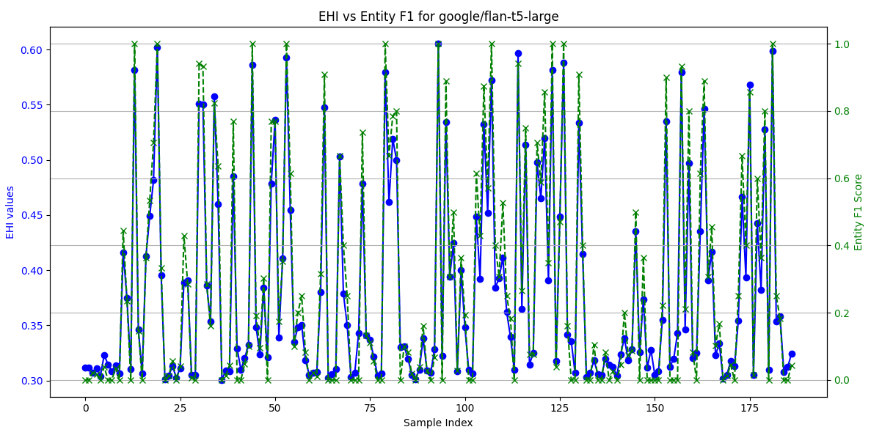}
\includegraphics[width=0.45\textwidth]{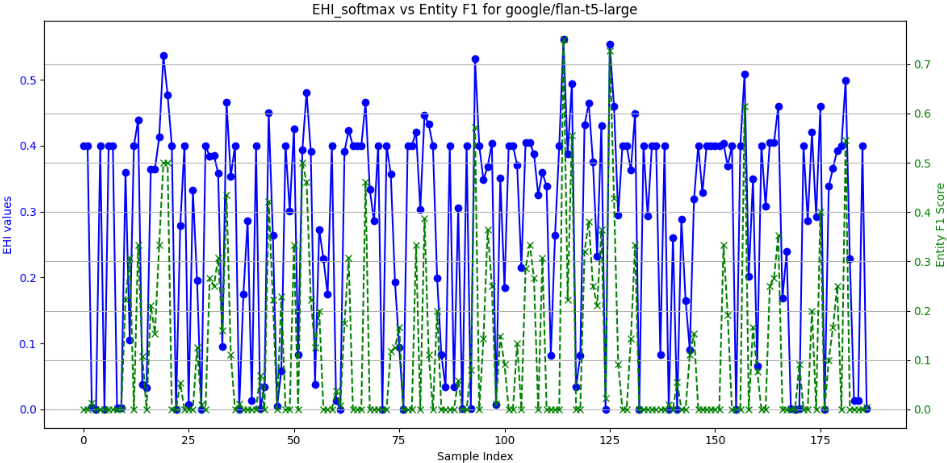}
\caption{Comparison of EHI and Entity F1 before (left) and after (right) EHI-guided fine-tuning for \texttt{google/flan-t5-large}. Left: EHI and F1 are volatile and weakly correlated. Right: EHI and F1 stabilize with a clearer inverse correlation.}
\label{fig:ehi-f1-comparison}
\end{figure*}

\section{Results}
\subsection{Task A: Summarization}

We evaluate the impact of EHI-guided fine-tuning on both hallucination control and entity prediction accuracy. Table~\ref{tab:results} summarizes the comparative results before and after fine-tuning.

The results of our summarization experiments show that the employed transformer-based models were able to generate summaries of moderate to high quality across different meeting transcripts. Each model had varying levels of success in capturing key points and reducing irrelevant content.

Initially, the generated summaries showed instances of factual inconsistencies, especially with respect to named entities and relational information. This led to the use of our proposed hallucination evaluation metrics --- Entity Hallucination Index (EHI)  --- which provided deeper insight into the factual alignment of generated content.

After applying domain-specific fine-tuning using the ELITR Minuting Corpus, we observed a notable qualitative improvement in summary generation. The summaries became:
\begin{itemize}
    \item More aligned with the actual meeting minutes,
    \item Less prone to hallucinated entities or relationships,
    \item Structurally more concise and contextually richer.
\end{itemize}

The improvement can be attributed to better adaptation of the model to the task-specific discourse patterns and vocabulary present in meeting transcripts. Fine-tuning helped the model focus on core informative content while minimizing hallucinated information. This suggests that even lightweight models, when fine-tuned effectively, can significantly improve their factual reliability. Moreover, the proposed hallucination metrics proved effective in detecting and guiding the correction of factual errors in summaries.

\subsubsection{Qualitative Examples with EHI Scores}

Table~\ref{tab:examples} presents sample summaries before and after reinforcement learning, along with their EHI breakdown. We observe that the fine-tuned model avoids inserting hallucinated entities and better preserves entity grounding.

\begin{table*}[t]
\centering
\begin{tabular}{p{0.45\textwidth} | p{0.45\textwidth}}
\toprule
\textbf{Before RL (EHI = 0.12)} & \textbf{After RL (EHI = 0.52)} \\
\midrule
The speaker discussed a new deal between IBM and Oracle to expand into Asian markets. & The speaker discussed the existing agreement with Oracle and Microsoft about service expansion. \\
\bottomrule
\end{tabular}
\caption{Example summaries before and after RL fine-tuning. The pre-RL summary includes a hallucinated entity ("IBM") not grounded in the transcript.}
\label{tab:examples}
\end{table*}

Further improvements can be achieved by:
\begin{itemize}
    \item Utilizing more sophisticated language models(e.g., LLMs like GPT-4, FLAN-T5 XXL, or domain-specific variants),
    \item Incorporating multi-modal signals (e.g., speaker roles, dialogue acts),
    \item Leveraging reinforcement learning or contrastive learning frameworks to penalize hallucinations during training.
\end{itemize}

\subsection{Task B: Question Answering}
Speaker identifiers were retained during Task B processing, as they can provide important contextual cues for identifying the source of factual information in the transcript.

\begin{table*}[t]
\centering
\begin{tabular}{l c}
\toprule
\textbf{Approach} & \textbf{Mean Score} \\
\midrule
baseline\_gpt-4o-2024-11-20 & 7.74 \\
baseline\_llama-3.1-8B-instruct & 7.08 \\
baseline\_phi-4-mini-instruct & 6.84 \\
baseline\_phi-3-small-128k-instruct & 6.65 \\
baseline\_llama-3.2-3B-instruct & 6.33 \\
GETALP@AutoMin & 4.55 \\
GETALP@AutoMin\_amr & 4.34 \\
GETALP@AutoMin\_amr\_only & 2.73 \\
\textbf{HallucinationIndexes@AutoMin (ours)} & \textbf{2.28} \\
\bottomrule
\end{tabular}
\caption{Mean scores of systems on the Monolingual (English-only) summarization task in AutoMin 2025.}
\label{tab:mono-task-results}
\end{table*}

For Task B, we implemented a QA system based on pre-trained transformer models and evaluated its performance across multiple meeting scenarios. The system was able to answer a variety of factual and reasoning-based questions using the provided transcripts.

While the base model performed well on direct factual queries, its performance occasionally dropped in cases involving long-range dependencies or contextual ambiguity. However, overall performance was satisfactory for a zero-shot setup. A key observation is that improvements in summary quality (Task A) can also contribute to more efficient QA systems by helping condense large transcripts into manageable, high-information inputs. Although our QA system did not explicitly rely on summaries, the synergy between summarization and QA was evident in terms of context clarity and answer precision.

The table \ref{tab:mono-task-results} depicts the results. Note that Automin evaluation considered various baselines, however the model Flan-T5-Large baseline was not considered. The scores in the evaluation had two limitations: (1)Note that Automin evaulation considered various baselines, however the model Flan-T5-Large baseline was not considered. (2) The hallucination index improves the quality focusing merely on entities and a subsequent method of Relation Hallucination is needed to complement it.

\subsection{Entity Hallucination Index (EHI) Improvements}

Before fine-tuning, EHI scores exhibited significant volatility, with frequent drops to 0.0, indicating frequent hallucinations. After fine-tuning, EHI scores became more consistent, largely stabilizing between 0.3 and 0.6, suggesting improved control over hallucinated entities.  The 

\subsection{Entity F1 Score Improvements}

Entity F1 scores improved markedly after fine-tuning. While initial F1 scores were often below 0.5 with many zeros (indicating missed entity matches), fine-tuning led to a surge in F1 scores, with numerous samples achieving values close to 1.0, reflecting high precision and recall for entity prediction.

\subsection{Correlation Between EHI and Entity F1}

Prior to fine-tuning, the relationship between EHI and F1 was weak or inconsistent. Post fine-tuning, we observe a stronger inverse correlation, indicating that as entity hallucinations decreased (higher EHI), entity prediction accuracy (F1) increased.
Figure~\ref{fig:ehi-f1-comparison} provides a visual comparison of EHI and Entity F1 scores across samples. Before fine-tuning, both metrics fluctuate significantly, indicating unstable entity prediction and hallucination control. After fine-tuning, the curves are more stable and exhibit clearer inverse correlation, validating the effectiveness of EHI-guided optimization.

\section{Discussion}

Beyond quantitative improvements, we conduct qualitative analysis to better understand how EHI-guided fine-tuning impacts summary generation.

\paragraph{Improved Entity Grounding.}
Before fine-tuning, generated summaries frequently introduced entities unrelated to the input transcript, or omitted key entities entirely. After fine-tuning, summaries exhibited better alignment with the input, correctly preserving mentioned organizations, speaker names, and events. We observe that entity mentions became more precise and contextually appropriate.

\paragraph{Enhanced Consistency Across Summaries.}
Fine-tuned models enhances EHI score which helps consistent entity coverage across samples. Hallucination behavior, which was previously erratic and dataset-dependent, stabilized substantially after training with EHI rewards as explained in \ref{tab:examples}. Overall summaries demonstrated less variance in entity recall and fewer abrupt entity shifts.

\paragraph{Limitations and Error Cases.}
While overall hallucination was reduced, occasional errors persisted, especially for rare or ambiguous entity mentions. In some cases, the model overly prioritized exact entity copying at the expense of paraphrasing or abstraction quality. This suggests a trade-off between strict entity grounding and higher-level semantic fluency that merits further investigation. This is where there is a need for Relation Hallucination detection and handling which the EHI is lacking.

\paragraph{Interpretation of EHI Improvements.}
The use of exponential scaling in the EHI formula emphasizes the balance between positive extraction and penalizing hallucination, which contributed to stable training dynamics. Moreover, EHI has a scope to be used readily in reference free context by combining with Relative Abstractiveness and generating EHI scores on reference data itself and use it as threshold for zero shot reference less summaries to predict hallucination.

\section{Conclusion}

In this work, we proposed a reward-driven fine-tuning approach for abstractive summarization, leveraging the Entity Hallucination Index (EHI) to directly optimize for entity faithfulness. By integrating EHI as a lightweight, reference-free reward signal, we reduced hallucinations and improved entity prediction accuracy without relying on human factuality annotations.

Experiments on meeting transcript datasets demonstrate that EHI-guided fine-tuning leads to more stable hallucination behavior and significant gains in entity F1 scores. Our results suggest that fine-grained entity-focused rewards provide an effective pathway for enhancing summarization reliability.

In future work, we plan to extend EHI-based fine-tuning to multi-document summarization and explore its integration with controllable generation frameworks.

\section*{Acknowledgements}
We would like to thank Naman Kabadi for his valuable support and assistance during the lab experiments related to this work.

\bibliography{custom}

\begin{thebibliography}{17}
\providecommand{\natexlab}[1]{#1}

\bibitem[{Cao et~al.(2020)Cao, Wang, and Wan}]{cao2020factual}
Shuyang Cao, Lu~Wang, and Xiaojun Wan. 2020.
\newblock Factual error correction for abstractive summarization models.
\newblock In \emph{Proceedings of the 2020 Conference on Empirical Methods in Natural Language Processing (EMNLP)}.

\bibitem[{Chung et~al.(2022)Chung, Hou, Longpre, Zoph, Tay, Fedus, Huang, Dai et~al.}]{chung2022scaling}
Hyung~Won Chung, Le~Hou, Shayne Longpre, Barret Zoph, Yi~Tay, William Fedus, Xuezhi Huang, Andrew~M. Dai, and 1 others. 2022.
\newblock Scaling instruction-finetuned language models.
\newblock \emph{arXiv preprint arXiv:2210.11416}.

\bibitem[{Durmus et~al.(2020)Durmus, He, and Diab}]{durmus2020feqa}
Esin Durmus, He~He, and Mona Diab. 2020.
\newblock Feqa: A question answering evaluation framework for faithfulness assessment in abstractive summarization.
\newblock In \emph{Proceedings of the 2020 Conference on Empirical Methods in Natural Language Processing (EMNLP)}.

\bibitem[{Falke et~al.(2019)Falke, Ribeiro, and Gurevych}]{falke2019ranking}
Tobias Falke, Leonardo F.~R. Ribeiro, and Iryna Gurevych. 2019.
\newblock Ranking generated summaries by correctness: An interesting but hard task.
\newblock In \emph{Proceedings of the 57th Annual Meeting of the Association for Computational Linguistics}.

\bibitem[{Goyal et~al.(2022)Goyal, Durmus, and Cardie}]{goyal2022fact}
Tirthankar Goyal, Esin Durmus, and Claire Cardie. 2022.
\newblock Fact-checking summarization with self-supervised and few-shot learning.
\newblock In \emph{Proceedings of the 60th Annual Meeting of the Association for Computational Linguistics (ACL)}.

\bibitem[{Honnibal and Montani(2017)}]{spacy2}
Matthew Honnibal and Ines Montani. 2017.
\newblock spacy 2: Natural language understanding with bloom embeddings, convolutional neural networks and incremental parsing.
\newblock \url{https://spacy.io}.

\bibitem[{Kingma and Ba(2015)}]{kingma2015adam}
Diederik~P Kingma and Jimmy Ba. 2015.
\newblock Adam: A method for stochastic optimization.
\newblock \emph{International Conference on Learning Representations (ICLR)}.

\bibitem[{Lin(2004)}]{lin-2004-rouge}
Chin-Yew Lin. 2004.
\newblock \href {https://aclanthology.org/W04-1013/} {{ROUGE}: A package for automatic evaluation of summaries}.
\newblock In \emph{Text Summarization Branches Out}, pages 74--81, Barcelona, Spain. Association for Computational Linguistics.

\bibitem[{Maynez et~al.(2020)Maynez, Narayan, Bohnet, and McDonald}]{maynez2020faithfulness}
Joshua Maynez, Shashi Narayan, Bernd Bohnet, and Ryan McDonald. 2020.
\newblock On faithfulness and factuality in abstractive summarization.
\newblock In \emph{Proceedings of the 58th Annual Meeting of the Association for Computational Linguistics}.

\bibitem[{Ouyang et~al.(2022)Ouyang, Wu, Jiang, Almeida, Wainwright, Mishkin, Zhang, Agarwal, Slama, Ray et~al.}]{ouyang2022training}
Long Ouyang, Jeff Wu, Xu~Jiang, Diogo Almeida, Carroll Wainwright, Pamela Mishkin, Chong Zhang, Sandhini Agarwal, Sandrine Slama, Alex Ray, and 1 others. 2022.
\newblock Training language models to follow instructions with human feedback.
\newblock In \emph{Advances in Neural Information Processing Systems (NeurIPS)}.

\bibitem[{Pagnoni et~al.(2021)Pagnoni, Padmakumar, and May}]{pagnoni2021understanding}
Artidoro Pagnoni, Vishakh Padmakumar, and Jonathan May. 2021.
\newblock Understanding factuality in abstractive summarization.
\newblock In \emph{Proceedings of the 2021 Conference of the North American Chapter of the Association for Computational Linguistics (NAACL)}.

\bibitem[{Pasunuru and Bansal(2018)}]{pasunuru2018multi}
Ramakanth Pasunuru and Mohit Bansal. 2018.
\newblock Multi-reward reinforced summarization with saliency and entailment.
\newblock In \emph{Proceedings of the 2018 Conference of the North American Chapter of the Association for Computational Linguistics (NAACL)}.

\bibitem[{Scialom et~al.(2021)Scialom, Dray, Lamprier, Piwowarski, and Staiano}]{scialom2021questeval}
Thomas Scialom, Paul-Alexis Dray, Sylvain Lamprier, Benjamin Piwowarski, and Jacopo Staiano. 2021.
\newblock Questeval: Summarization asks for fact-based evaluation.
\newblock In \emph{Proceedings of the 2021 Conference of the North American Chapter of the Association for Computational Linguistics (NAACL)}.

\bibitem[{Thonet et~al.(2024)Thonet, Rozen, and Besacier}]{thonet2024elitrbench}
Thibaut Thonet, Jos Rozen, and Laurent Besacier. 2024.
\newblock \href {https://arxiv.org/abs/2403.20262} {{ELITR-Bench: A Meeting Assistant Benchmark for Long-Context Language Models}}.
\newblock \emph{arXiv:2403.20262}.

\bibitem[{Williams(1992)}]{williams1992simple}
Ronald~J Williams. 1992.
\newblock Simple statistical gradient-following algorithms for connectionist reinforcement learning.
\newblock \emph{Machine learning}, 8(3-4):229--256.

\bibitem[{Yuan et~al.(2021)Yuan, Neubig, and Liu}]{10.5555/3540261.3542349}
Weizhe Yuan, Graham Neubig, and Pengfei Liu. 2021.
\newblock Bartscore: evaluating generated text as text generation.
\newblock In \emph{Proceedings of the 35th International Conference on Neural Information Processing Systems}, NIPS '21, Red Hook, NY, USA. Curran Associates Inc.

\bibitem[{Zhang* et~al.(2020)Zhang*, Kishore*, Wu*, Weinberger, and Artzi}]{Zhang*2020BERTScore:}
Tianyi Zhang*, Varsha Kishore*, Felix Wu*, Kilian~Q. Weinberger, and Yoav Artzi. 2020.
\newblock \href {https://openreview.net/forum?id=SkeHuCVFDr} {Bertscore: Evaluating text generation with bert}.
\newblock In \emph{International Conference on Learning Representations}.

\end{thebibliography}

\end{document}